# A Threshold-based Scheme for Reinforcement Learning in Neural Networks

Thomas H. Ward

thomas.holland.ward@gmail.com

## Abstract

*A generic and scalable Reinforcement Learning scheme for Artificial Neural Networks is presented, providing a general purpose learning machine. By reference to a node threshold three features are described 1) A mechanism for Primary Reinforcement, capable of solving linearly inseparable problems 2) The learning scheme is extended to include a mechanism for Conditioned Reinforcement, capable of forming long term strategy 3) The learning scheme is modified to use a threshold-based deep learning algorithm, providing a robust and biologically inspired alternative to backpropagation. The scheme may be used for supervised as well as unsupervised training regimes.*

## 1 Introduction

This paper proposes that a general purpose learning machine can be achieved by implementing Reinforcement Learning in an Artificial Neural Network (ANN), three interdependent methods which attempt to emulate the core mechanisms of that process are presented. Ultimately the biological plausibility of this scheme may be validated by reference to natural organisms. However that does not preclude the possibility that there is more than one underlying mechanism providing Reinforcement Learning in nature.

AI research has characteristically followed a bottom-up approach; focusing on subsystems that address distinct, specialized and unrelated problem domains. In contrast the work presented follows a distinctly top-down approach attempting to model intelligence as a whole system; a causal agent interacting with the environment [6]. The agent is not designed to solve a particular problem, but is instead assigned a reward condition. The reward condition serves as a goal, and in the path a variety of unknown challenges may be present. To solve these problems efficiently the agent requires intelligence.

This top-down approach assumes that the core self organizing mechanisms of learning that exist in natural organisms can be replicated in artificial autonomous agents. These can then be scaled up by endowing the agent with more resources (sensors, neurons & motors). Given sufficient resources and learning opportunities an agent may provide an efficient solution to a problem provided one exists. Also given the generalization properties of ANN's the agent can provide appropriate responses to novel stimuli.

A distinction is made between supervised, unsupervised and reinforcement training regimes. Supervised learning regimes use a (human) trainer to assign desired input-output pattern pairings. Unsupervised training regimes are typically used to cluster a data set into related groups. Reinforcement Learning (RL) may be considered a subtype of unsupervised training; it is sometimes called learning with a critic rather than learning with a teacher as the feedback is evaluative (right or wrong) rather than instructive (where a desired output action is prescribed). Significant RL successes have been achieved with the use of Temporal Difference (TD) methods [5][7], notably Q-learning[2].

First a definition of intelligence is required:

*The demonstration of beneficial behaviors acquired through learning.*

A beneficial action/behavior being one that would result in a positive survival outcome (eg successful feeding, mating, self preservation) for the agent. For the most part our inherent internal reward systems encourage us to perform beneficial behaviors, but this is not always the case (eg substance abuse may be rewarding but not beneficial). The term 'desirable behavior' is avoided due to existing usage of the term 'desired output' in supervised learning schemes.

Let's revise our definition, and expectation, of intelligence:

*The demonstration of rewarding behaviors acquired through learning.*

Rewarding behaviors/actions will be selected for reinforcement (ie learnt) over non rewarding ones. Rewarding behaviors are those that allow the agent to achieve the reward condition, thereby achieving its goal(s) in an acceptably efficient manner (eg elapsed time, steps taken, energy expended). Rewarding behaviors may lead to pleasure, or at least a reduction in pain. Goals are attained by achieving the pre-established reward condition, and thereby satiating active desire(s). Behaviors need not be active they may be passive; inaction may lead to reward and therefore be reinforced.

$$s_t \xrightarrow{a_t} s_{t+1}$$

From initial state $s_t$ if action $a_t$ results in an immediate reward in the subsequent state $s_{t+1}$, action $a_t$ will be reinforced. If the same (or similar from generalization) input pattern is encountered the learnt action will be performed. This process of learning is termed Primary Reinforcement. Primary Reinforcement reward conditions (eg hunger or thirst) typically drive some form of homeostatic, adaptive control function for the agent [5][8].

$$s_t \xrightarrow{a_t} s_{t+1} \xrightarrow{a_{t+1}} s_{t+2}$$

This is in contrast to Secondary/Conditioned Reinforcement where from initial state $s_t$ action $a_t$ does not result in immediate reward in subsequent state $s_{t+1}$. If later action $a_{t+1}$ does result in a reward in state $s_{t+2}$, this will lead to reinforcement of actions $a_{t+1}$ and $a_t$. The number of actions learnt from start to goal is arbitrary and depends on the relative size of the reward relative to the cost (or pain) in attaining it.

A learning scheme is presented that enables an embodied neural network agent to autonomously determine and learn desirable behaviors. The agent may be embodied in a real or artificial environment. Artificial environments may be modelled on physical environments or even abstract problem domains. The environment may be any set of input patterns, however there must be a causal relationship between the output (behavior) of the agent and the subsequent input pattern.

The common theme of this work is the application of thresholds to neural network activation functions, which thereby inform the learning process. Whilst the use of thresholds is by no means novel, mainstream learning methods are not heavily reliant upon them. By contrast the three presented features are all strictly dependent on the presence of a threshold.

Thresholds for neuron activation are widely found in animal cells, where sufficient 'excitation' is required to result in an electrical action potential or *'spike'* that can be signalled to other neurons [11]. Activations exceeding the threshold represent the occurrence of one or more spikes, with stronger activations representing sustained *'spike trains'* (fig 1).

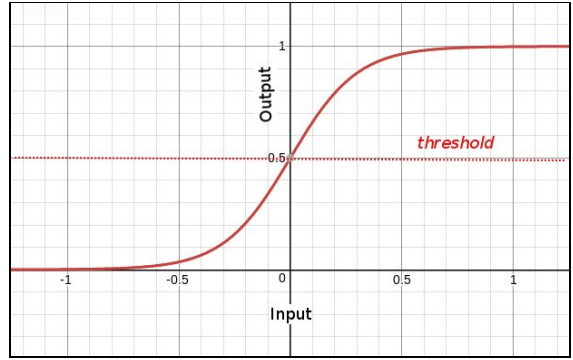

fig1 Logistic Activation Function with threshold

This paper presents three features:

1. A Primary Reinforcement learning scheme is achieved by wrapping a (supervised) backpropagation network within an unsupervised framework.

   *Primary Reinforcement enables 'desirable' actions (behaviors) to be autonomously generated and learnt; this is of particular benefit when a human supervisor is not available / does not know what the desired output should be, or when the environment, or agent itself, is changeable.*

2. The framework is then extended to provide Conditioned (Secondary) Reinforcement.

   *Conditioned (Secondary) Reinforcement enables an arbitrarily long sequence of chained behaviors to be autonomously learnt, in expectation of a primary reward; this provides long term strategy.*

3. An algorithm is described, termed Threshold Assignment of Connections (TAC), that replaces backpropagation within the framework, conversely this algorithm can also be used in supervised training schemes.

   *The Threshold Assignment of Connections algorithm provides a biologically inspired alternative to backpropagation.*

The examples that follow are focused mainly on target tracking, however this approach may be applied to a wide range of real and abstract problem spaces. The tasks are small in scale and primarily serve as proof of concept. While learning rate performance has been documented, the scheme has not been performance tuned. The intent of this work is to establish a biologically plausible working model of intelligence that is simple, scalable and generic.

## 2 Primary Reinforcement

In this section a learning scheme is described that provides Primary Reinforcement learning in an Artificial Neural Network. Weight adjustments using backpropagation[3] are traditionally used in supervised learning schemes; that is desired activations (or training sets) are established prior to the learning phase. In this example backpropagation will be used in an unsupervised learning scheme; desired actions will be generated on-the-fly. This approach, termed **Threshold Assignment of Patterns (TAP)**, relies on a threshold in the output layer to determine the desired output pattern.

Unlike mainstream methods, the presented scheme will create it's own actions *de novo* rather than relying on a predefined set. This provides a neural explanation of action selection and enables neural adaption should the causal relationship between the agent and environment alter (eg damaged motors, icy surfaces).

Although reward results in reinforcement the scheme is also stochastic; punishment results in new candidate behaviors being randomly generated. Sensing utilizes a sensory input pattern and behavior arises from a motor output pattern. Thinking (or processing) is implemented via layers of artificial neurons.

The learning scheme consists of the following components:

- Artificial Neural Network
- Learning Algorithm
- Framework
- Environment

The Artificial Neural Network architecture is that of a familiar multilayer perceptron (MLP)[3]. An input (sensor) pattern representing the environment, produces activations that feed forward and result in an output (motor) pattern representing a behavior.

The Learning Algorithm used to derive weight updates is the standard (supervised) back propagation learning algorithm[3][10].

The Framework sits between the network and environment. The framework acts as an interface between network and environment, and is responsible for establishing a reward condition as Primary Reinforcer and determining (desired output) behavior based on that reward condition. The framework, network and learning algorithm constitute the agent.

The Environment may be real or simulated. Simulated environments consist of states, and physics rules which define the relationship between states. The physics rules take the current (t) network output and determine which input pattern is next (t+1) presented to the network.

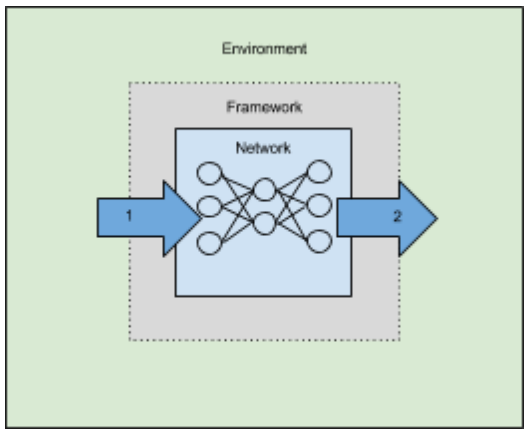
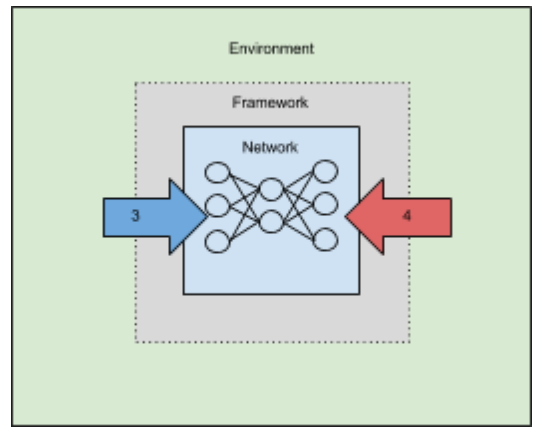

fig2 State 't'     fig3 State 't+1'

Backpropagation requires a set of desirable output patterns to be established. Through learning the desired output patterns will inform the network what the required output node activations are for each input pattern. In this scheme desired output patterns will be dynamically generated on-the-fly. But how can the agent determine potentially complex desired output patterns from a simple yes/no reward condition?

Desired activation values are set depending on whether a reward or punishment occurred after the output response. If a reward condition occurred all actual activations above threshold (eg fig 4 node 1) will be reinforced by setting the corresponding desired activation to 1.0 and all activations below threshold (eg fig 4 node 2) will be reinforced by setting the corresponding desired activation to 0.0. If punishment (no reward) occurred all actual activations, above or below threshold, will be weakened by setting desired activations to near threshold values.

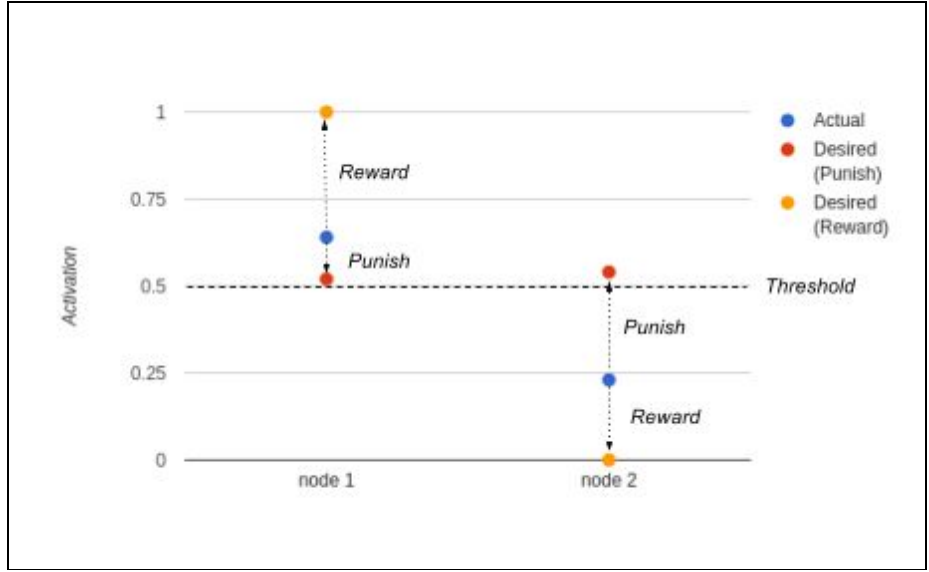

fig4   Desired activation : Punishment vs Reward

The desired activation $d_{u(t)}$ of an output layer node can be determined by reference to the actual activation of the reward node $a_{r(t+1)}$ in conjunction with the actual activation of the output layer node $a_{u(t)}$ (equation 1).

In reward conditions (reward threshold $\theta_r$ is achieved), the output node activation $a_{u(t)}$ will be strengthened; a maximal desired activation will be set if the output node threshold $\theta_u$ was achieved and a minimal value if it was not.

In punishment conditions (reward threshold $\theta_r$ is not achieved), the output node activation will be weakened by assigning it a random moderate desired activation, regardless of whether the output node threshold was achieved.

$$d_{u(t)} = \begin{cases} \begin{cases} 1 & \text{if } a_{u(t)} > \theta_u \\ 0 & \text{otherwise} \end{cases} & \text{if } a_{r(t+1)} > \theta_r \\ \{U[0.45..0.55] & \text{otherwise} \end{cases} \quad (1)$$

The agent will then learn by trial and error according to the following process (Box 1):

**Processing steps**

1. Input pattern for state 't' is presented and forward propagated through the network (Box 2). Node output activation may be in the range 0-1 (Box 3). A threshold value of 0.5 is assigned to each node in output layer.

2. Agent's behavior, based on whether output layer activations have exceeded threshold values (fig 1), determines subsequent input pattern 't+1'.

3. Framework determines 'reward' value based on 't+1' input pattern.

4. Weight changes are made:

    a. Desired 't' activations at the output layer are calculated:

        i. If reward condition then the desired 't' activation for that node is set to maximal value (1.0) for those that were above threshold, and set to minimal value (0.0) for those that were below threshold.

        ii. If no reward then the desired 't' activation for all nodes are random moderate values [0.45..0.55].

    b. Weight changes for state 't' are made according to error values which are back propagated through the network (Box 4).

Box 1 Threshold Assignment of Patterns processing steps

Input activation for unit u.

$$netinput_u = \sum_i weight_{ui}\, a_i$$

Box 2  Input activation

Output activation for unit u.

$$a_u = \frac{1}{1 + e^{-netinput_u}}$$

Box 3  Logistic Activation Function

1) Derive desired activation for output unit u.

$$d_{u(t)} = \begin{cases} \begin{cases} 1 & \text{if } a_{u(t)} > \theta_u \\ 0 & \text{otherwise} \end{cases} & \text{if } a_{r(t+1)} > \theta_r \\ U[0.45..0.55] & \text{otherwise} \end{cases}$$

2) Derive delta error value for an output layer node u, by finding difference between desired activation ($d_u$) and actual activation ($a_u$).

$$delta_u = (d_u - a_u) a_u (1 - a_u)$$

3) Derive weight change for connection between a hidden unit h and an output unit u, using learning rate.

$$\Delta weight_{uh} = lrate\ delta_u\ a_h$$

4) Derive delta error value for a hidden unit h, using weighted sum of all units in output layer.

$$delta_h = a_h (1 - a_h) \sum_u delta_u\ weight_{uh}$$

5) Derive weight change for connection between an input unit i and a hidden unit h, using learning rate.

$$\Delta weight_{hi} = lrate\ delta_h\ a_i$$

Box 4 Backpropagation weight update

In this way randomized desired out patterns will generate a new candidate behavior on the next presentation of that same (or similar) stimulus. In effect the response has been established before it is first manifested, and this rewarding response will be reinforced on future presentations. Conversely non-rewarding behaviors will be destroyed in favour of a new candidate behavior. Akin to natural selection, only rewarding behaviors will survive.

A behaviour (ie selected action) is represented by a generated output pattern of activity. Since output should be considered at motor rather than functional level, representations may be distributed rather than winner-take-all. The actual function of an action is determined by the causal relationship between the agent and environment. The output pattern will be determined by simultaneous excitation and inhibition arising from an input stimulus. In this sense the roles of *Sense-Think-Act* are tightly coupled.

A mapping is formed from an input pattern stimuli to an output pattern motor response, dependent on the reward that follows. No distinction is made between learning and testing phase; that is the agent in a continual process of learning and evaluation. A behavior is deemed to have been learnt when all output node activations are mature (eg above 0.9 or less than 0.1) for a given input pattern and results in a reward.

## 2.1 Example: Target tracking (part I)

### 2.1.1 Problem description

In this example the agent must autonomously learn how to track a target (fig 5).

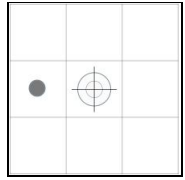

fig5 Target tracking example start state

- The target may appear in any one of nine cells of a grid (3x3). The agent is required to focus on the target, by moving it to the centre cell.
- The network is rewarded only if it moves the target to it's centremost cell. Once the target is moved to the centremost cell it is moved to a new starting position on the grid.
- The agent has no prior knowledge. It does not initially know it will be rewarded by relocating the target until it does so.
- The agent may only move the target 1 step (ie to an adjacent cell) in each iteration.
- To increase the task difficulty the agent may not move the target diagonally in one movement, two steps are required.

### 2.1.2 Network configuration

A network was created (table 1)(fig 6):

| Input layer | 9 input nodes; the centre node is designated as a special 'reward' node |
|---|---|
| Hidden layer | 12 hidden nodes |
| Output layer | 4 motor nodes |
| Notes | Each layer fully connected to the next via weights. Weights initially randomised. Each hidden and output nodes assigned an exclusive bias unit. Learning rate = 1.0 |

Table 1 Target tracking network configuration

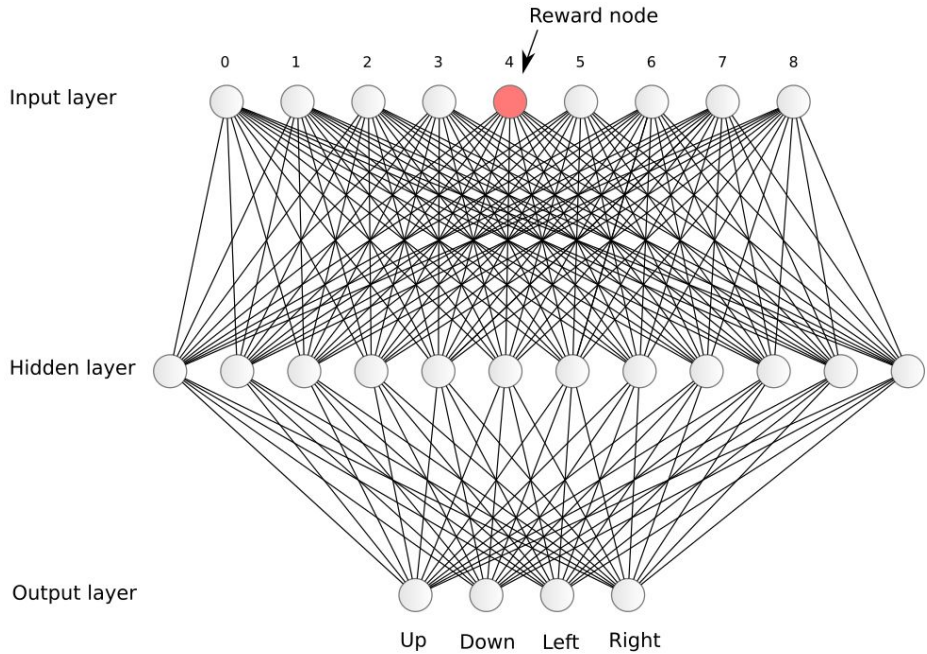

fig6 Target tracking network topology

The nine input nodes are mapped to each cell in the grid (fig7):

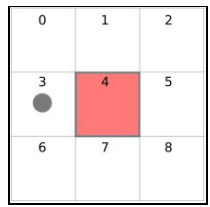

fig7 Target tracking input mapping

The network has four output nodes, representing output motors (Up, Down, Left, Right). Depending on the output the agent is able to move the target within the grid. In order for the agent to move the target in any given direction there must be only one output node activation above threshold (> 0.5), otherwise the target will remain stationary. Therefore there is only a 1-in-16 chance ($2^4$) the agent will move in the correct direction by chance.

The framework establishes the reward condition by assigning one the input nodes as a special 'reward node'. The framework consists of some rules to calculate previous (t) desired output patterns based on the current (t+1) input pattern which contains the reward value. A reward indicator value of 1 is set if the network it to be rewarded, and zero if it is to be punished. A reward occurs if the target is moved to the centre cell (fig 8):

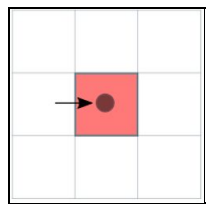

fig8 Target tracking reward condition

The behavior, or output response of the network, will causally influence the subsequent input pattern. Reward feedback will be given to the network, thereby informing whether the output was 'correct'. The output patterns are not predetermined. The network is presented a pattern and produces a response behavior. If the behavior was rewarding then a stronger version of the actual output is assigned as the desired output pattern. If the behavior was not rewarding a random pattern is set as the desired output pattern. A threshold is required at the output layer to make this decision.

*2.1.3 Results*

Initially the agent moves the target randomly. Activations tend to be weak (eg 0.4 - 0.6) across all output nodes. When rewarding behaviors are discovered these are further reinforced and the output matures (eg < 0.1, > 0.9).

With sufficient exposure the network learns the optimal behavior to achieve a reward; consistently moving one step left/right/up/down towards the food reward from starting locations (cells 1,3,5,7) (table 2). If the target is placed directly on the centre cell it will remain stationary.

| Scheme | # pattern presentations |
| --- | --- |
| Threshold Assignment of Patterns (Unsupervised backpropagation) | 116816 |

Table 2 Target tracking results

However, the agent is unable to learn how to move the target when placed in corners (cells 0,2,6,8). The physics rules prevent the agent from moving the target diagonally in one step, instead two steps are required. Whilst the agent was able to solve this '1-step' solution in 116816 presentations, it spent much of it's time temporarily 'stuck' in corner cells. The agent can only reinforce behaviors where there is an immediate reward in the subsequent input pattern. The network is unable to solve the 'temporal credit assignment' problem. In order to learn two or more consecutive behaviors secondary (conditioned) reinforcement is required (fig 9).

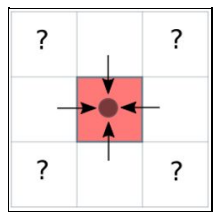

fig9 Primary Reinforcement partial solution for target tracking task

Note; In this example the reward condition was facilitated by assigning an existing input layer node as the 'Reward Node'. However, the reward condition could be evaluated against a combination of existing input layer nodes, a separate input layer node(s), or even no node at all (see XOR example below).

## 2.2 Example: XOR

*2.2.1 Problem description*

The agent will be required to solve the XOR problem (table 3) in order to test the network's ability to map non linear transformations that require multiple layers of neurons. This test will will also provide a performance comparison between unsupervised reinforcement learning and the supervised regime.

| (input) A | (input) B | (output) XOR |
|:---:|:---:|:---:|
| 0 | 0 | 0 |
| 0 | 1 | 1 |
| 1 | 0 | 1 |
| 1 | 1 | 0 |

Table 3 XOR task

Any input pattern can be considered an environment, and any output pattern a behavior. Thus any set of mappings can be learnt as they would under a conventional supervised learning regime. In contrast to the previous experiment, the network will now receive controlled exposure to all input patterns in turn. The behavior, or output response of the network will not influence the subsequent input pattern. However, reward feedback will be given to the network thereby informing whether the output was correct according to the desired pattern in the supervised training set. This tightly controlled presentation of input patterns is termed 'guided'.

For clarity, and to allow for a close comparison with the supervised backpropagation training regime, no explicit reward node will be established in the network topology. The framework is still responsible for setting the reward condition. To allow exposure to all the input patterns they will be cycled through in sequence. The framework will evaluate the output and decide whether it should be reinforced or not.

### 2.2.2 Network configuration

A network was created (table 4)(fig 10):

| | |
|:---:|:---|
| Input layer | 2 input nodes |
| Hidden layer | 3 hidden nodes |
| Output layer | 1 output node |
| Notes | Each layer fully connected to the next via weights. Weights initially randomised. Each hidden and output nodes assigned an exclusive bias unit. Learning rate = 1.0 |

Table 4 XOR network configuration

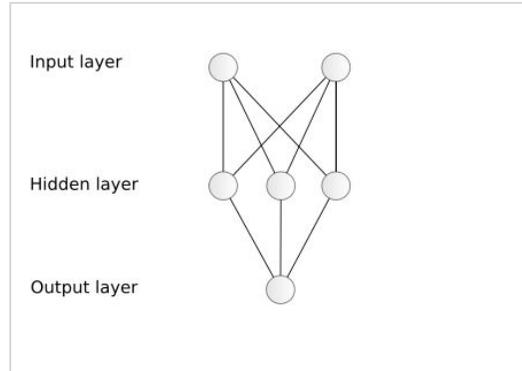

fig10    XOR network topology

### 2.2.3 Results

| Scheme | # pattern presentations |
|---|---|
| Supervised backpropagation | 2182 |
| Threshold Assignment of Patterns (unsupervised backpropagation) | 7550 |

Table 5 XOR results

The Threshold Assignment of Patterns (unsupervised backpropagation) regime required significantly more pattern presentations to learn the XOR solution than the supervised regime (table 5). Both were using the same set of initial weights and hyperparameters (learning rate etc). Both are using the same backpropagation algorithm to perform weight updates. The difference in performance can be attributed to the manner in which desired output patterns are provided. In the supervised scheme the desired patterns are known *a priori*, in the unsupervised (guided) learning regime these must be explored by the network through trial and error. This performance gap would be expected to widen as the number of nodes in the output layer grows and with it the number of candidate combinations.

## 2.3 Summary

Unsupervised Primary Reinforcement can be achieved, with reference to a threshold, by dynamically generating desired output activations and feeding these into a supervised learning algorithm. These experiments demonstrate the network's ability to learn and map arbitrary sets of patterns by reward, thereby producing behaviors that allow it to reach it's goal in an efficient manner. Primary Reinforcement may solve problems that are that are linearly inseparable. It is capable of training weights deep within the network, thereby capable of forming complex abstract representations.

The input patterns can be presented in any order. However, in order for the network to identify which output patterns (or behaviors) were correct, the network must be presented with subsequent reward feedback. The unsupervised Reinforcement Learning regime requires more iterations to learn compared to the supervised regime. The principal reason being that the unsupervised agent must explore the 'correct' solution through trial and error. And even when agent does not receive reinforcement, the new random candidate behavior may be a repeat of a prior incorrect one.

Primary Reinforcement may only learn a single behavior sequence, therefore it is not suitable for acquiring long term strategy that lead to distal rewards. Problems requiring long term strategy must use Secondary/Conditioned Reinforcement.

## 3 Secondary/Conditioned Reinforcement

Primary reinforcement provides a generic method of autonomously establishing beneficial output responses. But it has a significant limitation; it can only provide a one step mapping from start state(s) to goal. A more useful feature is the ability to establish an efficient series of behavior steps leading from start state(s) to a goal. This is the benefit provided by Conditioned Reinforcement. The term 'Conditioned' Reinforcement is preferred over that of 'Secondary' Reinforcement, since the latter may imply chained behaviors only 2 steps deep. In fact the number of chained behaviors can be arbitrarily deep, depending on the strength of the reward and subsequent reduction (discount factor) applied to it. This approach, termed **Threshold Assignment of Rewards (TAR)**, relies on building an association between a rewarding stimulus and an internal proxy reward.

In Primary Reinforcement a mapping is established between an input pattern stimuli to an output pattern motor response. In Secondary Reinforcement a mapping is established between an input pattern stimuli to an output pattern motor response AND a reward node. With sufficient reinforcement the response activation on the reward node matures; the reward node is now available as a proxy reward condition (secondary/conditioned reinforcer) for a given input pattern. In turn this conditioned reinforcer can help to create further conditioned reinforcers, that are activated only in response to a recognized input pattern stimuli.

The first conditioned response to be learnt will be closest to the primary reinforcer. Thereafter a chain of conditioned reinforcers can be established; via backward induction a series of 'breadcrumbs' are laid out in reverse from the goal. Using this mechanism planned or goal oriented tasks can be solved 'model-free'. The mapped reward node provides a *state-value function* for the agent. This mechanic resembles action-value mappings derived from *sample-backups* in SARSA and Q-learning methods[5][2]. However, in the present scheme state-action mappings are dealt with separately (TAP), and can be readily decoupled should an *actor-critic* architecture be employed [8]. Also while TD methods achieve exploration via (ε-greedy) probability, in the presented scheme exploration occurs when the agent fails to obtain a reward that satisfies a threshold value. If the agent is pursuing a suboptimal policy (path) to the goal, the threshold can be raised until the optimal path is found. Risky exploration can therefore be avoided unless required.

This approach is intended to overcome the temporal 'hill climbing' limitation outlined in the previous example. It is based on the a similar architecture and learning rule as before but with one important addition: **a special reward node is added to the output layer.** Unlike other nodes in the output layer the special reward node is not a motor neuron. Once this becomes mature it behaves like an input reward node; deciding which behaviors should be learnt. Behaviors leading to a distant reward can be chained together.

The prior desired activation $d_{u(t)}$ of an output layer reward node can be determined by reference to the subsequent actual activation of the output layer reward node $a_{u(t+1)}$ (equation 2).

In reward conditions (reward threshold $\theta_u$ is achieved), the output reward node activation $a_{u(t)}$ will be assigned the product of the subsequent actual activation of the output layer reward node $a_{u(t+1)}$ and discount factor $\gamma$.

In punishment conditions (reward threshold $\theta_u$ is not achieved), the output reward node activation will be assigned a minimal value.

$$d_{u(t)} = \begin{cases} \gamma a_{u(t+1)} & \text{if } a_{u(t+1)} > \theta_u \\ 0 & \text{otherwise} \end{cases} \quad (2)$$

Essentially mapping are now formed between input patterns and rewards, rather than just input patterns and motor nodes (Box 5).

---

**Processing steps**

*Differences to the Primary Reinforcement process described previously are highlighted in bold.*

1. Input pattern for state 't' is presented and forward propagated through the network (Box 2). Node output activation may be in the range 0-1 (Box 3). A threshold value of 0.5 is assigned to each node in output layer.

2. Agent's behavior, based on whether output layer activations have exceeded threshold values (fig 1), determines subsequent input pattern 't+1'.

3. Framework determines 'reward' value based on 't+1' input pattern **and on activation of special output reward node exceeding reward threshold (eg > 0.8 )**.

4. Weight changes are made:

    a. Desired 't' activations at the output layer are calculated:

        i. If reward condition then the desired 't' activation for that node is set to maximal value (1.0) for those that were above threshold, and set to minimal value (0.0) for those that were below threshold. **Set the desired activation for the special output reward node to discount (eg 95%) of reward value.**

        ii. If no reward then the desired 't' activation for all nodes are random moderate values [0.45..0.55]. **Set the desired activation for the special output reward node to minimal value (0.0).**

    b. Weight changes for state 't' are made according to error values which are back propagated through the network (Box 6).

*Note: To achieve this both the current and previous activations must be stored. Weight changes are derived using back propagation on previous activations.*

---

Box 5 Threshold Assignment of Rewards processing steps

1) Derive desired activation for reward output unit u.

$$d_{u(t)} = \begin{cases} \gamma a_{u(t+1)} & \text{if } a_{u(t+1)} > \theta_u \\ 0 & \text{otherwise} \end{cases}$$

2) Derive delta error value for an output layer node u, by finding difference between desired activation ($d_u$) and actual activation ($a_u$).

$$delta_u = (d_u - a_u)\, a_u\, (1 - a_u)$$

3) Derive weight change for connection between a hidden unit h and an output unit u, using learning rate.

$$\Delta weight_{uh} = lrate\; delta_u\; a_h$$

4) Derive delta error value for a hidden unit h, using weighted sum of all units in output layer.

$$delta_h = a_h\, (1 - a_h) \sum_u delta_u\; weight_{uh}$$

5) Derive weight change for connection between an input unit i and a hidden unit h, using learning rate.

$$\Delta weight_{hi} = lrate\; delta_h\; a_i$$

Box 6   Backpropagation weight update

## 3.1 Example: Target tracking (part II)

The same problem as described in section 3.1.1 is revisited. In this example the agent is equipped with the resources required to facilitate Secondary (Conditioned) Reinforcement.

### 3.1.1 Problem description

In this example the agent must autonomously learn how to track a target (fig 11).

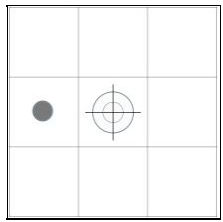

fig11    Target tracking example start state

- The target may appear in any one of nine cells of a grid (3x3). The agent is required to focus on the target, by moving it to the centre cell.
- The network is rewarded only if it moves the target to it's centremost cell. Once the target is moved to the centremost cell it is moved to a new starting position on the grid.
- The agent has no prior knowledge. It does not initially know it will be rewarded by relocating the target until it does so.
- The agent may only move the target 1 step (ie to an adjacent cell) in each iteration.
- To increase the task difficulty the agent may not move the target diagonally in one movement, two steps are

required.

## 3.1.2 Network configuration

A network was created (table 6)(fig 12):

| Input layer | 9 input nodes; the centre node is designated as a special 'reward' node |
|---|---|
| Hidden layer | 12 hidden nodes |
| Output layer | 4 motor nodes + 1 reward node |
| Notes | Each layer fully connected to the next via weights. Weights initially randomised. Each hidden and output nodes assigned an exclusive bias unit. Learning rate = 1.0 |

Table 6 Target tracking network configuration

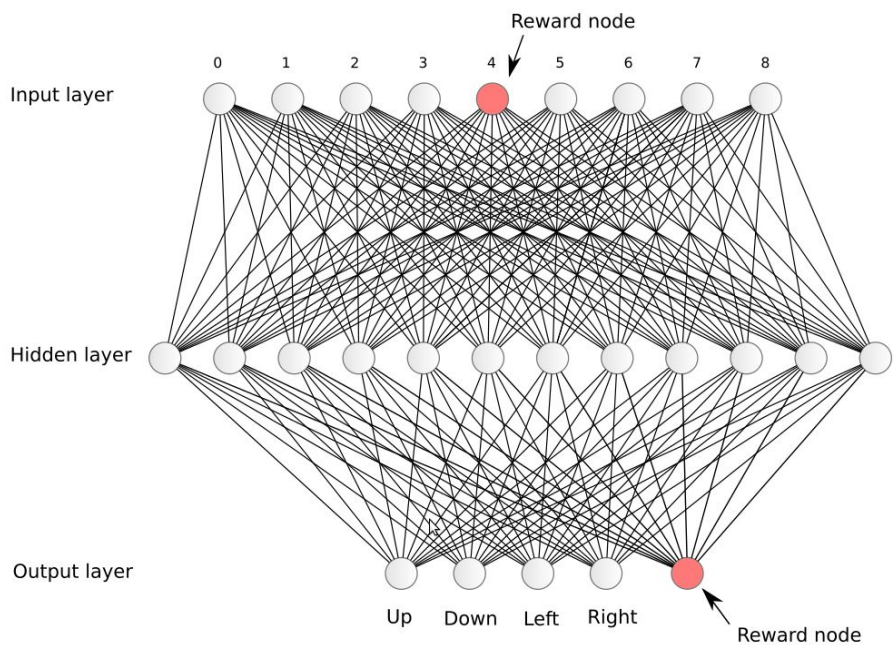

fig12

The nine input nodes are mapped to each cell in the grid (fig 13):

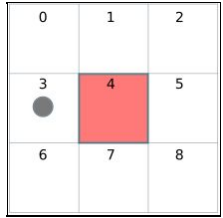

fig13     Target tracking input mapping

The network has four output nodes, representing output motors (Up, Down, Left, Right). Depending on the output the agent is able to move the target within the grid. In order for the agent to move the target in any given direction there must be only one output node activation above threshold (> 0.5), otherwise the target will remain stationary. Therefore there is only a 1-in-16 chance ($2^4$) the agent will move in the correct direction by chance.

The framework establishes the reward condition by assigning one the input nodes as a special 'reward node'. The framework consists of some rules to calculate previous (t) desired output patterns based on the current (t+1) input pattern which contains the reward value. A reward indicator value of 1 is set if the network it to be rewarded, and zero if it is to be punished. A reward occurs if the target is moved to the centre cell (fig 14):

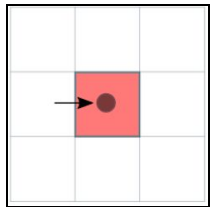

fig14     Target tracking reward condition

The network is also assigned an additional special reward node in the output layer. A reward also occurs if the activation of this output reward node is above the reward threshold (eg 0.8), which indicates it has been previously reinforced. If the subsequent input reward node or subsequent output reward node is activated, connections to the output reward node and motor output nodes will be reinforced.

Note; In this example a discrete special reward node was added to the output layer. It is possible to to have no dedicated reward node at output layer, instead it possible to test if activations of all motor neurons are high (and have therefore matured). Consequently activations, connections and therefore the strength of memories will be proportional to the reward.

### 3.1.3 Results

Initially the agent moves the target randomly. Activations tend to be weak (eg 0.4 - 0.6) across all output nodes. When rewarding behaviors are discovered these are further reinforced and the output matures (eg < 0.1, > 0.9).

| Scheme | # pattern presentations |
| --- | --- |
| Threshold Assignment of Reward (Unsupervised backpropagation) | 110324 |

Table 7 Target tracking results

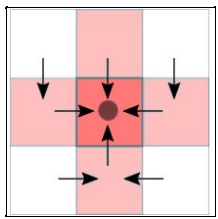

fig15    Conditioned Reinforcement full solution for target tracking task

The agent is able to learn how to move the target when placed in corners (cells 0,2,6,8) (table 7). The agent has effectively established proxy rewards in intermediate locations (cells 1,3,5,7) allowing chained sequences of behaviors to be learned (fig 15). The chaining of sequences of behaviors enables long term strategy to be acquired, this is demonstrated more substantially in the following maze navigation problem.

## 3.2 Example: Maze navigation

### 3.2.1 Problem description

In this example the agent must autonomously learn how to navigate a target through a maze (fig 16).

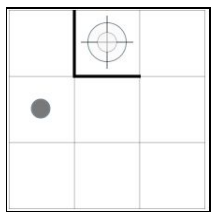

fig16    Maze task start state

- The target may appear in any one of nine cells of a grid (3x3). The agent is required to focus on the target, by moving it the top centre cell.
- Once the target is moved to the target cell (cell 1) it is moved back to it's original starting position on the grid (cell 3).
- The agent has no prior knowledge. It does not know that food will lead to a reward until it happens.
- The agent may only move the target 1 step (ie to an adjacent cell) in each iteration.
- To increase the task difficulty the agent may not move the target diagonally in one movement, two steps are required.
- To further increase task difficulty there is an invisible barrier between the start state and the goal. The agent must learn to navigate around the barrier.

### 3.2.2 Network configuration

A network was created (table 8) (fig 17):

| Input layer | 9 input nodes; the centre node is designated as a special 'reward' node |
|---|---|

| Hidden layer | 12 hidden nodes |
|---|---|
| Output layer | 4 motor nodes + 1 reward node |
| Notes | Each layer fully connected to the next via weights.<br>Weights initially randomised.<br>Each hidden and output nodes assigned an exclusive bias unit.<br>Learning rate = 1.0 |

Table 8 Maze task network configuration

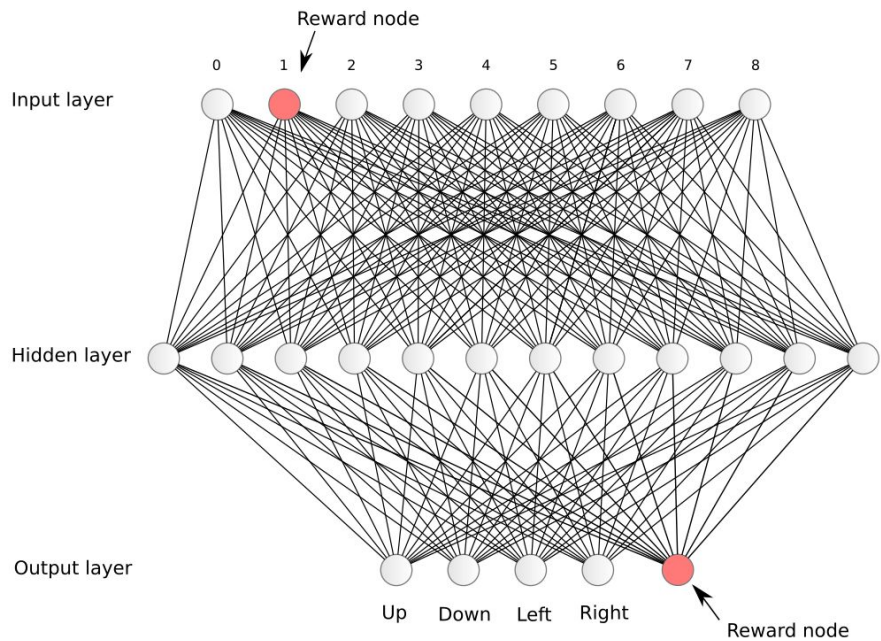

fig17    Maze task network topology

The nine input nodes are mapped to each cell in the grid (fig 18):

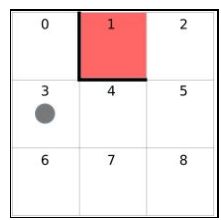

fig18    Maze task input mapping

The network has four output nodes, representing output motors (Up, Down, Left, Right). Depending on the output the agent is able to move the target within the grid. In order for the agent to move the target in any given direction there must be only one output node activation above threshold (> 0.5), otherwise the target will remain stationary. Therefore there is only a 1-in-16 chance ($2^4$) the agent will move in the correct direction by chance.

The framework establishes the reward condition by assigning one the input nodes as a special 'reward node'. The framework consists of some rules to calculate previous (t) desired output patterns based on the current (t+1) input pattern which contains the reward value. A reward indicator value of 1 is set if the network it to be rewarded, and zero if it is to be punished. A reward occurs if the target is moved to the top centre cell (cell 1).

The network is also assigned an additional special reward node in the output layer. A reward also occurs if the activation of this output reward node is above a given criteria (eg 0.8), which indicates it has been previously reinforced. Connections to the output reward node will be reinforced in the same manner as motor output nodes if the subsequent input reward node or subsequent output reward node is activated.

### 3.2.3 Results

Initially the network moves randomly through the environment. In time the agent is able to learn how to move the target around corners. The physics rules prevent the agent from moving the target diagonally in one step, instead two steps are required. The agent can only reinforce behaviors where there is an immediate reward in the next input pattern. The agent has effectively established proxy rewards, or sub-goals, in intermediate locations (cells 2,5,4) allowing chained sequences of behaviors to be learned (fig 19-22).

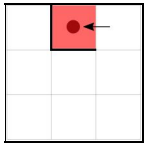
fig19 *Agent encounters the Primary Reinforcer and learns first behavior.*

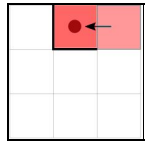
fig20 *Agent establishes first Conditioned Reinforcer.*

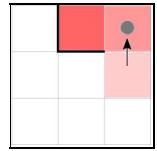
fig21 *Agent establishes a subsequent Conditioned Reinforcer by reference to the the initial Conditioned Reinforcer.*

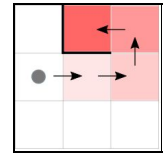
fig22 *Agent establishes a series of Conditioned Reinforcers from goal to starting position.*

With sufficient exposure the network learns the optimal behavior to achieve a reward; consistently moving one step left/right/up/down towards the goal from the starting location (cell 3).

The chaining of sequences of behaviors enables long term strategy to be acquired (table 9).

| Scheme | # pattern presentations |
|---|---|
| Threshold Assignment of Reward (Unsupervised backpropagation) | 80184 |

Table 9 Target tracking results

Learning rates can be improved by shaping [9], that is initially placing the agent closer to the reward and once learnt the distance may be increased.

## 3.3 Summary

A considerable challenge facing Reinforcement Learning schemes is that rewards can be very temporally delayed. Secondary/Conditioned Reinforcement provides an effective solution to this 'Temporal' Credit Assignment Problem. The reward condition was facilitated by adding single reward node to the output layer in addition to evaluating the reward at the input layer. The output reward desired activation experiences 'reduction' as it becomes temporally distant (number of iterations) from the target input reward. It is necessary to apply a discount factor to the output reward node in order to prevent the agent from becoming infinitely rewarded on a proxy reward that has been established. If the agent becomes fixated on a proxy reward it will become 'stuck', repeating the previous behavior.

The network learns appropriate mappings based on the temporal ordering of events; these mappings are contingent on subsequent reward conditions. Learning is based on cause and effect. It is nondeterministic; alternate policies may be used to attain a reward. The scheme does not rely on *eligibility traces* [1], or require long-term retention of prior network states (other than t-1 activations).

If the reward condition is removed the agent displays extinction; behaviours weaken and eventually become random.

It is also possible to achieve this effect without an explicit motivator node in output layer. Since only strong (output) behaviors are those which have been reinforced, a test can be made against the strength of the entire output pattern rather than a specific node.

When using a single node representation for the (mapped) output layer reward node and applying a discount factor, a diminishing desired activation is set for that node. The node is interpreted in an analogue (continuous) fashion. An alternative representation would be to introduce a discrete output reward node for each time step away from the input node reward.

# 4 Threshold Assignment of Connections

So far backpropagation, an algorithm intended for supervised learning, has been housed within an unsupervised framework; desired output patterns have been presented by a framework acting as a surrogate supervisor. Using the same framework as described in earlier sections, the backpropagation learning algorithm was replaced with an alternative algorithm, termed **Threshold Assignment of Connections (TAC)**.

A biological plausibility concern faces supervised learning schemes in general:

- How are desired output patterns selected and presented to the network ?

In previous examples, once backpropagation is placed in the Reinforcement Learning framework it is no longer a truly supervised learning scheme; desired output patterns are not known *a priori*. However backpropagation faces the additional biological plausibility concern:

- How is error back propagated ?

Thus far research has provided little neurobiological support for backpropagation [4][15]. TAC does not require backward connections or tenuous biochemical explanations. TAC could be explained by neuromodulators rather than at the neural computation level. This would provide a more robust solution with a simpler biological explanation.

The desired activation $d_{u(t)}$ of any node can be determined by reference to the actual activation of the reward node $a_{r(t+1)}$ in conjunction with the actual activation of the postsynaptic node $a_{u(t)}$ and presynaptic node $a_{h(t)}$ (equation 3).

If the presynaptic node did not fire connections from that node will not be strengthened or weakened; if the presynaptic node threshold $\theta_h$ was not achieved the desired activation will remain as the actual activation $a_{u(t)}$. If the presynaptic node did fire connections from that node may be strengthened or weakened.

In reward conditions (reward threshold $\theta_r$ is achieved), the postsynaptic node activation $a_{u(t)}$ will be strengthened; a maximal desired activation will be set if the postsynaptic node threshold $\theta_u$ was achieved and a minimal value if it was not.

In punishment conditions (reward threshold $\theta_r$ is not achieved), the postsynaptic node activation will be weakened by assigning it a random moderate desired activation, regardless of whether the postsynaptic node threshold was achieved.

$$d_{u(t)} = \begin{cases} \begin{cases} \begin{cases} 1 & \text{if } a_{u(t)} > \theta_u \\ 0 & \text{otherwise} \end{cases} & \text{if } a_{r(t+1)} > \theta_r \\ \{U[0.45..0.55] & \text{otherwise} \end{cases} & \text{if } a_{h(t)} > \theta_h \\ \{a_{u(t)} & \text{otherwise} \end{cases} \quad (3)$$

Direct 'on-node' delta values that can be used to calculate weight updates, rather than backpropagating them (Box 7).

---

**Processing steps**

*Differences to the Conditioned Reinforcement process described previously are highlighted in bold.*

1. Input pattern for state 't' is presented and forward propagated through the network (Box 2). Node output activation may be in the range 0-1 (Box 3). A threshold value of 0.5 is assigned **to all nodes in every layer.**

2. Agent's behavior, based on whether output layer activations have exceeded threshold values (fig 1), determines subsequent input pattern 't+1'.

3. Framework determines 'reward' value based on 't+1' input pattern **and on activation of special output reward node exceeding reward threshold (eg > 0.8 )**.

4. Weight changes are made:

    a. Desired 't' activations **on every node** are calculated:

        i. If reward condition then the desired 't' activation for that node is set to maximal value (1.0) for those that were above threshold, and set to minimal value (0.0) for those that were below threshold. **Set the desired activation for the special output reward node to 95% of reward value.**

        ii. If no reward then the desired 't' activation for all nodes are random moderate values [0.45..0.55]. **Set the desired activation for the special output reward node to minimal value (0.0).**

    b. **Weight changes for state 't' are made for all connections to each node in situ**

**according to error values (Box 8); these are not back propagated.**

*Note: To achieve this both the current and previous activations must be stored.*

Box 7 Threshold Assignment of Connections processing steps

1) Calculate desired activation for postsynaptic unit u.

$$d_{u(t)} = \begin{cases} \begin{cases} \begin{cases} 1 & \text{if } a_{u(t)} > \theta_u \\ 0 & \text{otherwise} \end{cases} & \text{if } a_{r(t+1)} > \theta_r \\ U[0.45..0.55] & \text{otherwise} \end{cases} & \text{if } a_{h(t)} > \theta_h \\ a_{u(t)} & \text{otherwise} \end{cases}$$

2) Derive delta error value for postsynaptic node u, by finding difference between desired activation ($d_u$) and actual activation ($a_u$).

$$delta_u = (d_u - a_u) \, a_u \, (1 - a_u)$$

3) Derive weight change for connection between presynaptic unit h and postsynaptic unit u, using learning rate.

$$\Delta weight_{uh} = lrate \, delta_u \, a_h$$

Box 8 Threshold Assignment of Connections weight change

## 4.1 Example: Maze navigation

The same problem as described in section 3.2.2 is revisited. In this example the agent is equipped with the TAC algorithm rather than backpropagation.

### *4.1.1 Problem description*

In this example the agent must autonomously learn how to navigate a target through a maze (fig 23).

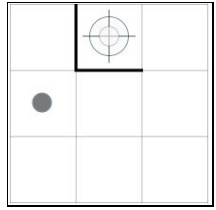

fig23    Maze task start state

- The target may appear in any one of nine cells of a grid (3x3). The agent is required to focus on the target, by moving it the top centre cell.
- Once the target is moved to the target cell (cell 1) it is moved back to it's original starting position on the grid (cell 3).
- The agent has no prior knowledge. It does not know that food will lead to a reward until it happens.
- The agent may only move the target 1 step (ie to an adjacent cell) in each iteration.
- To increase the task difficulty the agent may not move the target diagonally in one movement, two steps are required.
- To further increase task difficulty there is an invisible barrier between the start state and goal. The agent must learn to navigate around the barrier.

### 4.1.2 Network configuration

A network was created (table 10)(fig 23):

| Input layer | 9 input nodes; the centre node is designated as a special 'reward' node |
|---|---|
| Hidden layer | 12 hidden nodes |
| Output layer | 4 motor nodes + 1 reward node |
| Notes | Each layer fully connected to the next via weights. Weights initially randomised. Each hidden and output nodes assigned an exclusive bias unit. Learning rate = 1.0 |

Table 10 Maze task network configuration

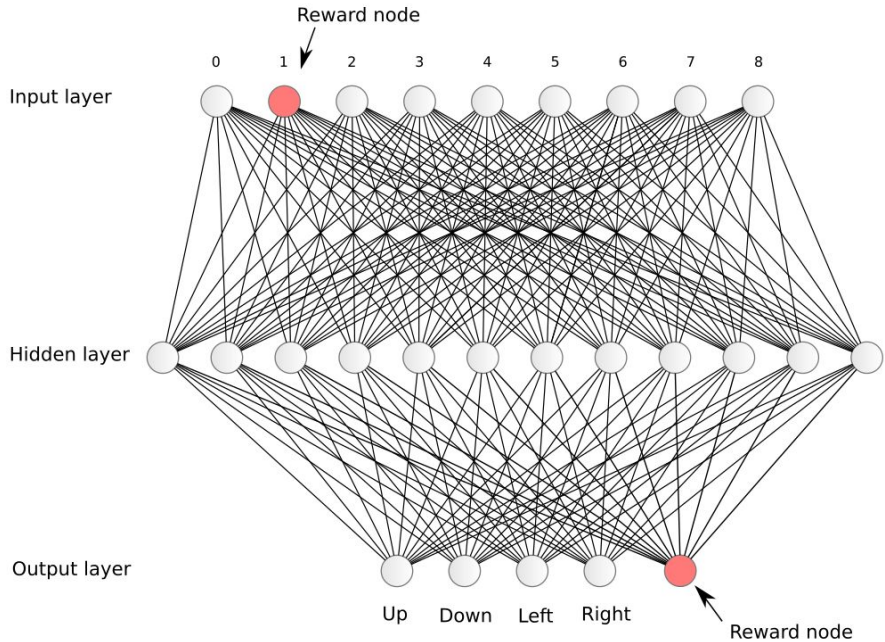

fig24    Maze task network topology

The nine input nodes are mapped to each cell in the grid (fig 25):

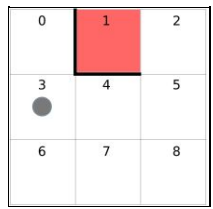

fig25    Maze task input mapping

The network has four output nodes, representing output motors (Up, Down, Left, Right). Depending on the output the agent is able to move the target within the grid. In order for the agent to move the target in any given direction there must be only one output node activation above threshold (> 0.5), otherwise the target will remain stationary. Therefore there is only a 1-in-16 chance ($2^4$) the agent will move in the correct direction by chance.

The framework establishes the reward condition by assigning one the input nodes as a special 'reward node'. The framework consists of some rules to calculate previous (t) desired output patterns based on the current (t+1) input pattern which contains the reward value. A reward indicator value of 1 is set if the network it to be rewarded, and zero if it is to be punished. A reward occurs if the target is moved to the top centre cell (cell 1).

The network is also assigned an additional special reward node in the output layer. A reward also occurs if the activation of this output reward node is above a given criteria (eg 0.8), which indicates it has been previously reinforced. If the subsequent input reward node or subsequent output reward node is activated, connections to the output reward node and motor output nodes will be reinforced.

### *4.1.3 Results*

Initially the network moves randomly through the environment. In time the agent is able to learn how to move around corners to the goal. The physics rules prevent the agent from moving the target diagonally in one step, instead two steps are required. The agent can only reinforce behaviors where there is an immediate reward in the next input pattern. The agent has effectively established proxy rewards, or sub-goals, in intermediate locations (cells 2,5,4) allowing chained sequences of behaviors to be learned (fig 26-29).

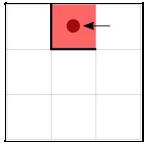
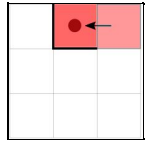
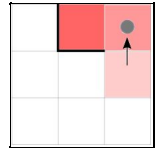
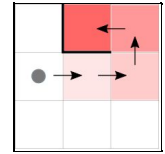

fig26  *Agent encounters the Primary Reinforcer and learns first behavior.*

fig27  *Agent establishes first Conditioned Reinforcer.*

fig28  *Agent establishes a subsequent Conditioned Reinforcer by reference to the the initial Conditioned Reinforcer.*

fig29  *Agent establishes a series of Conditioned Reinforcers from goal to starting position.*

With sufficient exposure the network learns the optimal behavior to achieve a reward; consistently moving one step left/right/up/down towards the food reward from the starting location (cell 3).

The chaining of sequences of behaviors enables long term strategy to be acquired (table 11).

| Scheme | # pattern presentations |
| --- | --- |
| Threshold Assignment of Reward (Unsupervised backpropagation) | 80184 |
| Threshold Assignment of Connections (Unsupervised) | 18212 |

Table 11 Maze task tracking results

Learning rates can be improved by shaping [9], that is initially placing the agent closer to the reward and once learnt increasing the distance.

## **4.2 Example: XOR**

The same problem as described in section 3.1.2 is revisited. In this example the agent is equipped with the TAC algorithm rather than backpropagation.

### *4.2.1 Problem description*

The agent will be required to solve the XOR problem (table 12) in order to test the network's ability to map non linear transformations that require multiple layers of neurons. This test will will also provide a performance comparison between unsupervised reinforcement learning and the supervised regime.

| (input) A | (input) B | (output) XOR |
|---|---|---|
| 0 | 0 | 0 |
| 0 | 1 | 1 |
| 1 | 0 | 1 |
| 1 | 1 | 0 |

Table 12 XOR task

The behavior, or output response of the network will not influence the subsequent input pattern. However, reward feedback will be given to the network thereby informing whether the output was correct. Once again the output patterns are not predetermined. This tightly controlled presentation of input patterns is termed 'guided'. The physics rules are modified in order to guide the agent.

### 4.2.2 Network configuration

A network was created (table 13) (fig 30):

| Input layer | 2 input nodes |
|---|---|
| Hidden layer | 3 hidden nodes |
| Output layer | 1 output node |
| Notes | Each layer fully connected to the next via weights. Weights initially randomised. Each hidden and output nodes assigned an exclusive bias unit. Learning rate = 1.0 |

Table 13 XOR Network configuration

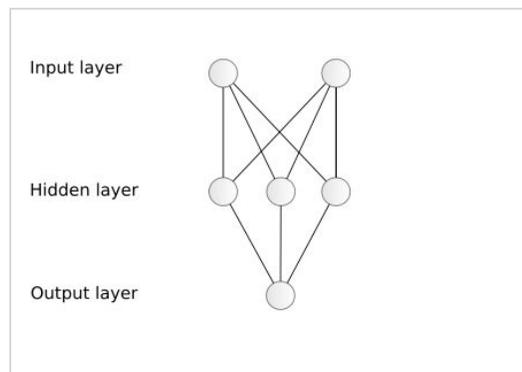

fig30    XOR topology

*4.2.3 Results*

| Scheme | # pattern presentations |
|---|---|
| Supervised backpropagation | 2182 |
| Threshold Assignment of Connections (unsupervised) | 311 |

Table 14 XOR results

The Threshold Assignment of Connections (unsupervised) scheme required significantly less pattern presentations to learn the XOR solution than the supervised regime (table 14). Both were using the same set of initial weights and learning parameters (learning rate etc). Despite the fact that in the supervised scheme the desired patterns are known *a priori*, while in the unsupervised (guided) learning regime these must be discovered by the network through trial and error. However this performance gap would be expected to lessen as the number of nodes in the output layer grows, increasing dimensionality and with it the number of candidate combinations.

## 4.3 Summary

These experiments demonstrate TAC may solve linearly inseparable problems that are normally reserved for supervised training regimes. It is capable of training weights deep within the network, thereby capable of forming complex abstract representations. TAC can also solve problems that require long term strategy, when applied in conjunction with Secondary (Conditioned) Reinforcement. Once TAC has replaced backpropagation, the framework is no longer required to act as a surrogate supervisor.

It should be noted that while the scheme suggests assigning random desired activations around the threshold value [0.45..0.55] in punishment conditions, it was found to be at least as effective to set this in the range [0-1]. It should be appreciated that (with moderate learning rates) this has the same effect of establishing an 'immature' response around the threshold level (fig 1). These alternatives have implications as to what similar mechanisms we might expect to find in natural organisms.

# 5 Discussion

The merits of the present scheme can be evaluated in terms of Machine Learning Applications as well as the broader Cognitive Science implications (eg biological, psychological, philosophical).

## 5.1 Applications

*5.1.1 Primary Reinforcement (TAP)*

One of the key advantages of Primary Reinforcement over supervised learning schemes is that the agent is able to autonomously explore solutions (behaviors) to problems, of use when a human may be unable to provide training data. Another benefit of Primary Reinforcement is that learning is dynamic and continuous, there is no distinction between training and classification phases. This is useful when the agent encounters novel stimuli, or when a once useful response is no longer so.

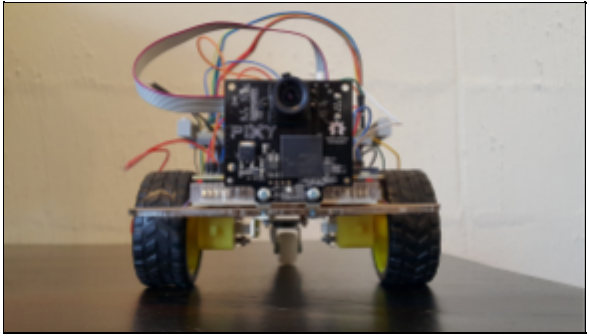

*'Rags the robot' learns how to track objects with a vision sensor. The robot has an onboard Arduino microcontroller and is wirelessly controlled by desktop computer.*

fig31     *'Rags the robot'*

The learning scheme can be readily applied to robotics. As proof of concept an agent was created to remotely control 'Rags the robot'; a mobile Arduino based device (fig 31). The task was to target an object, in this case a red frisbee. For this purpose Rags was equipped with a vision sensor (PixyCam), and was capable of rotating via a pair of motor controlled wheels. The agent begins each learning session with randomised network weights. Similar to the previous targeting tasks, the robot initially moves randomly, rotating passed the target. Eventually the agent is able to co-ordinate its motors and align itself with the target. A key benefit of TAP is that action is dynamically generated by the network itself, rather than being reliant on a predefined set of actions. If the causal relationship between agent and environment changes (eg rotation direction of the wheel motors is inverted), Rags will compensate via neural adaption.

The Primary Reinforcement solution presented is readily scalable; additional sensors, motors and hidden units can be easily added without reworking the underlying learning process. These additional resources will be utilized in problem solving. Alternate behaviors will be adopted in the face of unforeseen environmental hazards, or if the agent suffers sensor/motor damage.

Although Primary Reinforcement enables autonomous agents to discover solutions autonomously, a potential issue is that the learned behavior may not be ideal but is sufficient to achieve a reward condition; this can be termed a *suboptimal action (behavior) limitation*. For example a ball may have been successfully kicked into a goal, but the kicking technique itself was poor. To avoid this a higher standard can be achieved by setting a more stringent reward condition (eg lowering the reward value or raising the threshold), the tradeoff being more optimal behaviors are potentially slower to be explored and learnt.

Another limitation of the present solution is that we are unable to set arbitrary desired output activations on output nodes, a maximal or minimal value is set (ie 1.0 or 0.0). For example in a network with three output nodes we may choose a desired output pattern vector of [0.6, 0.3, 0.7] for a supervised network, but in the present scheme the desired output pattern vector would be limited to [1.0, 0.0, 1.0]. This shortcoming can be termed a *binary limitation*. To workaround this limitation alternate output representations may be required to achieve the same level of granularity.

One of the challenges of scaling relates to the complexity of output responses; the more output nodes that are required for a rewarding behavior to occur, the longer it will take for the agent to explore candidate behaviors and discover the requisite combination. Also it should also be noted that not all node activations may be relevant to producing a rewarding response. These may be included in the dynamically generated desired output pattern, and the agent is unable to differentiate which nodes were responsible for achieving the reward. These redundant activations may be eliminated by conforming to what can be termed a *laziness principle*, where rewards are reduced in proportion to output node activations.

Note; while Reinforcement Learning is essentially a trial and error approach to learning, it is possible to first part train in supervised mode, save the weights, and then switch to unsupervised mode.

### 5.1.2 Secondary reinforcement (TAR)

A potential issue facing the presented scheme may be termed a *suboptimal policy (path) limitation*. At first glance this is similar to the *suboptimal behavior limitation* described in Primary Reinforcement, but on a temporal level. There may be a tendency to *exploit* known rewards rather than *explore* new ones. The problem being that the while the acquired sequence of behaviors reliably leads to the goal, and the individual behaviors may have been optimal, the route was not the most efficient one possible. Once learnt the agent has no pressure to alter its preferred policy, a condition which afflicts natural organisms. An optimal policy can be found by decreasing the reward value or increasing the threshold relative to reward, but since exploration involves potential risk it can be avoided unless required. In addition behaviors may be *'shaped'* to elicit the optimal policy [9].

### 5.1.3 Threshold Assignment of Connections (TAC)

TAC is significantly easier to implement than backpropagation. It also provides a fully neural explanation of Reinforcement Learning, with processing essentially only at the node level. This would greatly simplify any hardware implementations of the method. It has fewer error derivation dependencies than backpropagation, and hence should be more resilient to damage/information loss occurring during weight updates. It is also is not expected to suffer from the vanishing gradient problem. However, further scalability testing is required for TAC (eg increasing network depth & breadth).

In terms of performance, TAC was found to learn in fewer iterations than supervised backpropagation. However this advantage is expected to diminish as the number of output nodes, and therefore potential candidate behaviors, increase.

## 5.2 Limitations

### 5.2.1 Timing issues and Actor-Critic architecture

In the examples provided the action function (TAP) is coupled with that of the value function (TAR), since they both share the same network architecture. Consequently the output for the two functions are derived simultaneously. This presents a timing issue, since activations are fed forward through the action component before it has been given the opportunity to learn from output of the value component.

This timing issue can be avoided by use of an actor-critic architecture [8][14], providing separate pathways for the two functions. The value function (TAR) can thereby be derived and used to dictate learning for itself and also for the action function (TAR) prior to new actions being issued.

This would have implications regarding how we would expect connection weights in these two pathways to respond to output from the value function. In the value pathway we would expect weights to be based on prior (t-1) activations (since new activations have passed through this pathway), but in the action pathway we would expect weight updates to be based on existing (t) activations (since new activations have not yet passed through this pathway).

## 5.3 Final words

With reference to a threshold an entirely neural based explanation of Reinforcement Learning has been presented. A threshold has been used to provide the roles of action generation and selection in Primary Reinforcement, the roles of exploitation and exploration in Conditioned Reinforcement, and the role of credit assignment in multi-layered networks. These roles are interdependent and rely on the causal interaction between agent and environment.

The scheme presented provides a self organizing model of cognition, being fundamentally hedonistic with learning driven by reward. The scheme is based on a single continuum of pleasure and pain, with a single learning mechanism for the two; pleasure effectively being the alleviation of pain. Whilst the examples provided are configured with only a single drive, this should be extended to multiple drives working in concert/competition. As a rule of thumb the agent should conform to a 'laziness principle', that is they should be punished for exerting unnecessary energy, thereby achieving their goal with greater efficiency. The examples provided entail the drive being always active, and therefore continually reinforcing or weakening behaviors. However, it should be taken that when not active (eg drives are satiated), behaviors will not be reinforced or weakened and the agent will not engage in active learning. This does not preclude the scheme being augmented with other forms of learning (eg Hebbian). Also while the examples presented are based on feedforward architectures the present scheme is not mutually exclusive with recurrency [13]. Indeed it is envisaged LSTM [12] and other methods of recurrency could be used to enhance the presented scheme, providing information of prior state.

The scheme's true potential lies in embodiment, whether in real or artificial environments. Agents may also be placed in abstract environments that do not physically exist (eg a stock market). The scheme is generic and readily scalable, and given sufficient time and resources may tackle a wide variety of tasks. Tasks can be assigned either by setting of the reward condition goal, or by placing in the path to that goal. If the presented scheme of Reinforcement Learning should be deemed biologically or psychologically implausible, the practical benefits remain of utility.

## 5.4 Appendix

*5.4.1 Platform*

Desktop PC (i3 3.7GHz, 16GB ram)

OS Linux Mint 17.3

Neural Network software written in C (GCC) : https://github.com/thward/neural_agent